# APPROACHES TO FRAUD DETECTION ON CREDIT CARD TRANSACTIONS USING ARTIFICIAL INTELLIGENCE METHODS

Yusuf Yazici

Department of Computer Engineering,
Istanbul Technical University, Istanbul, Turkey

## ABSTRACT

*Credit card fraud is an ongoing problem for almost all industries in the world, and it raises millions of dollars to the global economy each year. Therefore, there is a number of research either completed or proceeding in order to detect these kinds of frauds in the industry. These researches generally use rule-based or novel artificial intelligence approaches to find eligible solutions. The ultimate goal of this paper is to summarize state-of-the-art approaches to fraud detection using artificial intelligence and machine learning techniques. While summarizing, we will categorize the common problems such as imbalanced dataset, real time working scenarios, and feature engineering challenges that almost all research works encounter, and identify general approaches to solve them. The imbalanced dataset problem occurs because the number of legitimate transactions is much higher than the fraudulent ones whereas applying the right feature engineering is substantial as the features obtained from the industries are limited, and applying feature engineering methods and reforming the dataset is crucial. Also, adapting the detection system to real time scenarios is a challenge since the number of credit card transactions in a limited time period is very high. In addition, we will discuss how evaluation metrics and machine learning methods differentiate among each research.*

## KEYWORDS

*Credit Card, Fraud Detection, Machine Learning, Survey, Artificial Intelligence*

## 1. INTRODUCTION

The number of cashless transactions is at its peak point since the beginning of the digital era and it is most likely to increase in the future. While that is an advantage and provides ease of use for customers, it also creates opportunities for fraudsters. Only in 2016, 34,260.6 million transactions have been performed, making a total of 66,089 transactions per second. The net loss of the global economy out of fraudulent transactions is $2.17 billion [1].

As the loss is quite major, there is a number of research to decrease the causalities created by credit card fraud. The number of research papers in the finance application area using machine learning reaches to thousands [2]. While some of them try to solve this using mathematical rule-based algorithms, recently, machine learning and artificial intelligence techniques are in demand. That is the result of the big data collected from billions of transactions, and this data somehow could be useful in trying to predict whether a next, unknown transaction is actually a fraud or not. Credit card fraud could be categorized into two types: Application and behavioral fraud [3]. Application fraud could be defined as getting new cards from banks or companies using false or others' information. Behavioral fraud is, on the other hand, could be mail theft, stolen cards, counterfeit card or 'card holder not present fraud' [4].

Detecting a credit card fraud is in fact a binary classification problem, where the outcome is either false or true. That classification problem could be solved using three machine learning tasks: supervised learning, unsupervised learning, and semi-supervised learning. A supervised learning approach utilizes past, known transactions that are labeled as fraudulent or legitimate. The past data is trained and the created model is used to predict whether a new transaction is fraud or not. Unsupervised techniques are the ones that do not use the labeled data, but use unlabeled data to characterize the data distribution of transactions [6]. With that way, the outlier data could be acceptable as the fraudulent transactions. Clustering and compression algorithms are used to solve unsupervised problems. When the two approaches stated are combined, it brings out semi-supervised algorithms. These algorithms are generally used when there is less labeled data in the dataset.

The most common supervised learning algorithms are given below: Logistic Regression, Decision Trees, Neural Networks, Support Vector Machines (SVMs), Naïve Bayes, K-nearest neighbor (KNN).

In this survey, we will explore which state-of-the-art machine learning algorithms and methods are used to solve fraud detection after 2018 to present-day. Besides, while researches are carried out, there are some common problems or challenges almost all works encounter such as imbalanced dataset, doing the correct feature engineering, or adapting the work to real-time scenarios. We will categorize these challenges by examining state-of-the-art papers.

The rest of the paper is organized as follows: The approaches that are used to detect fraud detection on credit card transactions using machine learning methods are given in the Section 2. Section 3 desribes the common challenges while conducting these works. Lastly, Section 4 concludes the paper.

## 2. APPROACHES USING MACHINE LEARNING METHODS

Common algorithms and methods used in other works are presented in this section. Table 1. briefly depicts the used methods in the researches.

Akila et al. present an ensemble model named Risk Induced Bayesian Inference Bagging model, RIBIB [1]. They propose a three-step approach: a bagging architecture with a constrained bag creation method, Risk Induced Bayesian Inference method as base learner, and a weighted voting combiner. Bagging is a process of combining multiple training datasets, and utilizing them separately to train multiple classifier models [1]. They evaluated their solution on Brazilian Bank Data and exceled at cost minimizing compared to the other state-of-the-art models.

Sá et al. propose a customized Bayesian Network Classifier (BNC), that is automatically created by an algorithm named Hyper-Heuristic Evolutionary Algorithm (HHEA) [5]. HHEA builds a custom BNC algorithm, creating an ultimate combination of necessary modules for the dataset at hand. The dataset they used is UOL PagSegure, which is an online payment service in Brazil. They evaluated their results using $F_1$ score and a term they called as economic efficiency, measuring the company's economic loss from fraud. They also use an approach called instance reweighing while comparing their results with other baselines. It is basically reweighing (assigning them more importance) false negatives (predicted as legitimate but actually fraudulent), as they are more important for the payment companies (than the other way around) [5].

Carcillo et al. combine supervised and unsupervised techniques and presents a hybrid approach to fraud detection [6]. The reason for using unsupervised learning is that fraudulent behavior changes over time and the learner needs to consider these fraud patterns. They specify an approach to calculate outlier scores at different granularity levels. One of them is global granularity, in which all samples of transactions are considered in one global distribution. The other is local granularity, where outlier scores are computed independently for each credit card. Lastly, cluster granularity is in between global and local granularity, where customer behavior, such as amount of money spent at the last 24h is taken into account [6]. To implement their classifier model, they used Balanced Random Forest (BRF) algorithm. Topn Precision and AUC-PR (Area under the precision-recall curve) metrics are used for evaluation.

Table 1. Examining various works for fraud detection

| | Algorithm | Real dataset | Derived Features | Evaluation Metrics |
|---|---|---|---|---|
| Akila et al. (2018) | Risk Induced Bayesian Inference Bagging | √ | | Cost based, FPR, FNR, TNR, TPR, Recall, AUC |
| Sá et al. (2018) | HHEA based Bayesian Network Classifier | √ | | $F_1$, economic loss |
| Carcillo et al. (2019) | K-means, Balanced Random Forest | √ | | AUC-PR, Precision |
| Eunji et al. (2019) | Logistic regression, decision trees, recurrent neural networks, convolutional neural networks | √ | | K-S Statistics, AUROC, Alert rate, precision, recall, cost reduction rate |
| Sikdar et al. (2020) | Decision tree based Intuitionistic fuzzy logic | √ | √ | Sensitivity, specificity, false negative rate, false positive rate, precision accuracy and F-measure |
| Lucas et al. (2020) | Random Forest Classifier with derived features created from Hidden Markov Model (HMM) | √ | √ | Precision-Recall AUC |
| Misra et al. (2020) | Dimension reduction using Deep Autoencoders, Multi-Layer Perceptron, K-Nearest Neighbor, Logistic Regression | √ | | $F_1$ |
| Dornadula et al. (2019) | Decision Trees, Naïve Bayes classification, Least Squares Regression, Support Vector Machines (SVMs) | √ | √ | Accuracy, Precision, Matthews Correlation Coefficient (MCC) |
| Carta et al. (2019) | Prudential Multiple Consensus model, Ensemble of Multi-layer perceptron, Gaussian Naïve Bayes, Adaptive Boosting, Gradient Boosting, Random Forest | √ | | Sensitivity, Specificity, AUC, Miss rate, Fallout |
| Nami et al. (2018) | Dynamic Random Forest (DRF), with minimum risk model K-Nearest Neighbor (KNN) | √ | √ | Recall, Precision, F-measure Specificity, Accuracy |
| Wang et al. (2019) | Decision tree, Naïve Bayes, Logistic regression, Random Forest, and Artificial Neural Network | √ | | False negative, false positive rate |

Eunji et al. compare two different approaches to fraud detection: Hybrid ensemble methods and deep learning. They do this comparison in a framework named champion-challenger analysis [7]. The champion model is a model that is used for a while that uses various machine learning classifiers such as decision trees, logistic regressions, and simple neural network. Each method is trained using different samples and features and the best outcome is chosen manually by experts [7]. Whereas the challenger framework uses recent deep learning architecture consisting of convolutional neural networks, recurrent neural networks and their variants. This framework tries each modern deep learning architecture, specifies activation functions, dropout rates and costs, then finds the best hyperparameters. It then uses early stopping, a way to stop training when no further improvement is achieved on the validation set, during training. It finally chooses the best performed model and saves the hyperparameters and complexity settings used on them and tries to find better candidates out of previous experiments [7].

Various evaluation metrics are used to compare each framework: K-S statistics: Maximum value of difference between two distributions, AUROC: Area under receiver operating characteristics, plot of true positive rate over false positive rate. Alert rate: given a cut-off value to alert by the user, transactions are alerted over all transactions. Precision: fraudulent transactions predicted as true (TP) over all alerted transactions (TP + FP). Recall: fraudulent transactions are predicted as true (TP) over fraudulent transactions (TP + FN). Cost reduction rate: the missed frauds (FN) cost the transaction amount to the owner company. It is calculated using the sum of the amount that came out as FNs. As a result, the challenger framework which is based on deep learning performs much better than the champion framework.

Sikdar et al. developed a decision tree using intuitionistic fuzzy logic. They argue that it recognizes the conceptual properties of attributes of transactions, so that legitimate ones are not captured as fraud, or vice-versa [8]. The motivation under using fuzzy logic is that it is not tried as much as the other artificial intelligence methods on e-transactional fraud detection.

C4.5 algorithm is used with fuzzy logic and intuitionistic fuzzy logic and the final algorithm is named IFDTC4.5. The fuzzy tests are defined by different attributes and the information gain ratio is calculated using membership degree and non-membership degree. That information is then used to create an intuitionistic fuzzy logic algorithm that classifies between fraud, normal, and doubtful transactions. To evaluate the final model, almost all suitable metrics are used: Sensitivity, specificity, false negative rate, false positive rate, precision accuracy and F-measure. They show that the proposed method outperforms the existing techniques. Also, this algorithm is argued to work more efficiently and fast compared to others [8].

Pourhabibi et al. review graph-based anomaly detections between 2007-2018. They declare that the general approach is to do the right feature engineering and graph embedding into a feature space, so that the machine learning models could be built [9]. They also argue that graph-based anomaly detection techniques have been on the raise since 2017 [9].

As it could be guessed, credit card transactions are not independent events that are isolated; instead, they are a sequence of transactions [10]. Lucas et al. take this property into account and create Hidden Markov Model (HMM) to map a current transaction to its previous transactions, extract derived features, and use those features to come up with a Random Forest classifier for fraud detection [10]. The features created by HMM quantify how similar a sequence is to a past sequence of a cardholder or terminal [10]. They evaluate the final model using Precision-Recall AUC metric and showed that feature engineering with HMM presents an acceptable rise in the PR-AUC score.

Misra et al. propose a two-stage model for credit card fraud detection. First, an autoencoder is used to reduce the dimensions so that the transaction attributes are transformed into a lower dimension feature vector [11]. Then, the final feature vector is sent to a supervised classifier as an input. An autoencoder is a type of a feed-forward neural network. It regularly has the same input and output dimensions, yet there exists a reconstruction phase in-between. Initially, there is an encoder that transforms the input to a lower dimension, then the encoder's output tries to construct the output layer with the same dimension as the input layer. That step is performed by the decoder. In this work, only the encoder part of the autoencoder is used. Subsequently, the output from the encoder is used as an input to a number of classifiers: Multi-Layer perceptron, k-nearest neighbors, logistic regression. $F_1$ score is used to evaluate the final classifier [11]. It outperforms similar methods in terms of $F_1$.

Dornadula et al. discuss that card transactions are frequently not similar to the past transactions made by the same cardholder [12]. Therefore, they first group the cardholders based on their transaction amount: High, medium, and low range partitions. Afterwards, they extract some extra features based on these groups using the sliding-window method [12]. Next, SMOTE (Synthetic Minority Over-Sampling Technique) operation is performed on the dataset to solve the imbalance dataset problem. Precision and MCC (Matthews Correlation Coefficient) measures are used to evaluate the model. Among various classifiers, logistic regression, decision tree and random forest models perform well based on the evaluation metrics.

Carta et al. consolidate state-of-the-art classification algorithms with a model called Prudential Multiple Consensus. The idea is built upon the fact that the results of different classifiers are not the same in terms of certain transactions [13]. The algorithm is formed of two steps:

1) A transaction is perceived as legitimate if and only if the current algorithm classifies it as legitimate and the classification probability is above the average of all algorithms. Otherwise, it is perceived as fraudulent.

2) Majority voting is applied after all algorithms run the first step and the final decision is made [13].

Sensitivity, fallout, and AUC evaluation metrics are used to evaluate the model in terms of combination of a number of algorithms such as Multi-layer perceptron, Gaussian Naïve Bayes, Adaptive Boosting, Gradient Boosting, and Random Forest. It performs well in terms of Sensitivity and AUC.

Nami et al. present a two-stage solution to the problem. Before starting the algorithm steps, they derive some extra features to acquire an enhanced understanding of cardholders' spending behavior [14]. Then, at the first stage, reasoning that new attitudes of cardholders would be closer to their recent attitudes, a new similarity measure is constructed based on transaction time. That measure naturally designates more weight to recent transactions [14]. The second stage consists of training a Dynamic Random Forest algorithm applying a minimum risk model. It is a model to decide the outcome of a transaction with a cost-sensitive approach [15]. Nami et al. tested their model using various metrics such as recall, precision, f-measure specificity, and accuracy and showed that the minimum risk approach made an increase in performance.

Wang et al. combine machine learning algorithms with customer incentives [22]. They argue that there must be secondary verification in order to achieve more accurate results. The secondary verification could be applied to certain transactions that are higher than a threshold value. They specify the strategies and their conditions according to the benefits they offer to retailers, card issuers, and consumers, resulting in a "win-win-win" success [22]. The existing strategies generally are no-prevention (doing nothing), and using machine learning techniques for all transactions. The third strategy is to make a second verification with customer incentives [22]. They experiment the different strategies with algorithms Decision tree, Naïve Bayes, Logistic regression, Random Forest, and Artificial Neural Network.

## 3. COMMON CHALLENGES

The credit card fraud detection problem shares some common challenges to consider while implementing efficient machine learning algorithms: They could be grouped as overcoming imbalanced dataset problem, doing the right feature engineering, and executing models in real-time scenarios.

### 3.1. Imbalanced Dataset Problem

Almost all datasets of banks or other organizations contain millions of transactions, and all of them share a common problem in terms of state-of-the-art machine learning algorithms: Imbalanced dataset. The problem arises from the fact that the rate of actual fraud transactions out of all transactions is nominal. The number of legitimate transactions per day in 2017 completed by Tier-1 issuers is 5.7m, whereas fraud transactions in the same category is 1150 [17]. This unbalanced data distribution lessens the effectiveness of machine learning models [18]. Hence, training models to detect fraudulent transactions that are very nominal requires extra caution and thinking. The general known approaches to the imbalanced dataset problem are categorized into two: Sampling methods, and cost-based methods [30]. We examine how state-of-the-art research tackles this problem.

Fiore et al. solve the problem by increasing the number of "interesting but underrepresented" instances in the training set [16]. They achieve this by generating credible examples using Generative Adversarial Networks (GANs) that mimics "interesting" class examples as much as possible [16]. From the point of view of sensitivity rate, the classifier generated by the help of GANs gives sufficient results compared to the original classifier.

Rtayli et al. state that the quantity of transactions that are fraudulent is a very small portion of total transactions, and that brings out the imbalanced dataset problem. To solve that issue, they use Random Forest Classifier to select only relevant features [24]. They use this approach in the area of Credit Risk Identification and it gives accurate results based on the metrics they use; this approach could also be used in the credit card fraud detection.

Zeager et al. state that there are common approaches to overcome class imbalance which are oversampling the minatory class (fraudulent transactions), undersampling the majority class (legitimate transactions), and cost-sensitive cost functions. They utilize an oversampling approach named SMOTE (synthetic minority oversampling technique), that generates synthetic examples of fraudulent transactions [26].

Jurgovsky et al., on the other hand, present a different approach to [26] and employ and undersampling an account level to overcome class imbalance [27]. In depth, they tag accounts that contain at least one fraudulent transaction as "compromised", and tag accounts that do not contain any fraudulent transactions as "genuine". With a probability of 0.9, they randomly pick a genuine account, and pick a compromised account with a probability of 0.1. The process is repeated times to create a training set [27].

Zhu et al. suggest an approach called Weighted Extreme Learning Machine (WELM) to solve imbalanced dataset problems [28]. WELM is a transformed version of ELM for imbalanced datasets assigning different weights to different types of samples [28].

## 3.2. Feature Engineering Challenge

The pure transaction information extracted from the organization database is quite restricted. The balance of cardholder, transaction time, credit limit, transaction amount are some of them. When only these ready-to-use features are used to train common machine learning algorithms, the performance is not likely to vary among them. In order to create a difference, accurate feature engineering becomes a must. We will look through some research that handle feature engineering in credit card fraud detection scenarios.

Zhang et al. generate a feature engineering method that is dependent on homogeneity-oriented behavior analysis, stating that behavior analysis should be done on distinct groups of transactions with the same transaction characteristics [19]. These characteristics could be extracted from the information of time, geographic space, transaction amount, and transaction frequency. For each characteristic found, two strategies are processed for feature engineering: Transaction aggregation and rule-based strategy [19].

Roy et al. extend the baseline features and add the following new features to their model [20]: Frequency of transactions per month, filling the missing data dummy variables, maximum, mean authorization amounts in the 8-month period, new variables to indicate when a transaction is made at a predefined location such as restaurants, gas stations etc., a new variable demonstrating whether a transaction amount in a given retailer is greater than 10% of the standard deviation of the mean of legitimate transactions for that retailer [20].

Chouikh et al. utilize Convolutional Neural Networks in fraud detection analysis and they argue that since deep learning algorithms use deep architecture internally, they extract their features automatically in a hierarchical way with layers navigating from bottom to upwards. With that way, a feature engineering process that is time and resource consuming is avoided [21].

Wu et al. are interested in credit cards rather than transactions for feature engineering [25]. They mainly focus on the credit card cash-out problem. It is a fraud technique that spends all limit on the credit card. The study incorporates additional features into the model by receiving information from industry experts, tips shared by fraudsters online, reports, and news. The number of total features the study reaches is 521, creating a pool for feature selection studies [25]. The classifier model created using these feature sets increases the precision performance by 4.6%-8.1% [25].

## 3.3. Real Time Working Problems

Since the incoming transactions that are processed to the system every day are excessive, and behaviors of cardholders and fraudsters could change in a rapid way, the classification models should be regenerated frequently. This exposes the question of how efficient the created models are. We investigate some research done that tries to implement efficient systems to work in a real-time manner.

Carcillo et al. utilize open source big data tools such as Apache Spark, Kafka and Cassandra to create a real-time fraud detector called Scalable Real-time Fraud Finder (SCARFF) [23]. They emphasize that the system is tested extensively in terms of scalability, sensitivity, and accuracy;

and it processes 200 transactions per second, which they argue that is much more than their partner, with a rate of 2.4 transactions per second [23].

Patil et al. suggest a fraud detection system on credit cards on a real-time basis analyzing incoming transactions. It uses Hadoop network to encode data in HDFS format and the SAS system converts the file to raw data. The raw data is transferred to the analytical model in order to build the data model. That cycle helps the system learn the model by itself in a scalable and real-time manner [29].

## 4. CONCLUSIONS

As the digital era matures, the number of transactions that are processed with credit card rises continuously. Fraudsters could abuse that rise and could convert a possible advantage into a disadvantage. Research is continued to be conducted for how to detect these kind of transactions. This survey presented how machine learning and artificial intelligence methods are utilized to detect credit card frauds. Subsequently, some common challenges such as imbalanced dataset, feature engineering, and real-time working scenarios, that are encountered during the progress are examined by the research basis. It could be concluded that the path is not over to adapt a machine learning fraud detection system to real-time environment since most of the works conducted still prefer an offline detection mechanism and the number of resarch for solving the problem is relatively low. On the other hand, the improvement on managing imbalanced dataset and feature engineering challenges is apparent. There exist some operative and effective methods to solve each problem and they considerably increase the model performances.

Future work can be performed to improve real-time scenarios combined with sufficient feature engineering and state-of-the-art machine learning methods.

## REFERENCES

[1]  S. Akila and U. Srinivasulu Reddy, "Cost-sensitive Risk Induced Bayesian Inference Bagging (RIBIB) for credit card fraud detection," Journal of Computational Science, vol. 27, pp. 247–254, Jul. 2018, doi: 10.1016/j.jocs.2018.06.009.
[2]  A. M. Ozbayoglu, M. U. Gudelek, and O. B. Sezer, "Deep learning for financial applications : A survey," Applied Soft Computing, vol. 93, p. 106384, Aug. 2020, doi: 10.1016/j.asoc.2020.106384.
[3]  Y. Jin, R. M. Rejesus *, and B. B. Little, "Binary choice models for rare events data: a crop insurance fraud application," Applied Economics, vol. 37, no. 7, pp. 841–848, Apr. 2005, doi: 10.1080/0003684042000337433.
[4]  S. Bhattacharyya, S. Jha, K. Tharakunnel, and J. C. Westland, "Data mining for credit card fraud: A comparative study," Decision Support Systems, vol. 50, no. 3, pp. 602–613, Feb. 2011, doi: 10.1016/j.dss.2010.08.008.
[5]  A. G. C. de Sá, A. C. M. Pereira, and G. L. Pappa, "A customized classification algorithm for credit card fraud detection," Engineering Applications of Artificial Intelligence, vol. 72, pp. 21–29, Jun. 2018, doi: 10.1016/j.engappai.2018.03.011.
[6]  F. Carcillo, Y.-A. Le Borgne, O. Caelen, Y. Kessaci, F. Oblé, and G. Bontempi, "Combining unsupervised and supervised learning in credit card fraud detection," Information Sciences, May 2019, doi: 10.1016/j.ins.2019.05.042.
[7]  E. Kim et al., "Champion-challenger analysis for credit card fraud detection: Hybrid ensemble and deep learning," Expert Systems with Applications, vol. 128, pp. 214–224, Aug. 2019, doi: 10.1016/j.eswa.2019.03.042.
[8]  S. M. S. Askari and M. A. Hussain, "IFDTC4.5: Intuitionistic fuzzy logic based decision tree for E-transactional fraud detection," Journal of Information Security and Applications, vol. 52, p. 102469, Jun. 2020, doi: 10.1016/j.jisa.2020.102469.

[9] T. Pourhabibi, K.-L. Ong, B. H. Kam, and Y. L. Boo, "Fraud detection: A systematic literature review of graph-based anomaly detection approaches," Decision Support Systems, vol. 133, p. 113303, Jun. 2020, doi: 10.1016/j.dss.2020.113303.

[10] Y. Lucas et al., "Towards automated feature engineering for credit card fraud detection using multi-perspective HMMs," Future Generation Computer Systems, vol. 102, pp. 393–402, Jan. 2020, doi: 10.1016/j.future.2019.08.029.

[11] S. Misra, S. Thakur, M. Ghosh, and S. K. Saha, "An Autoencoder Based Model for Detecting Fraudulent Credit Card Transaction," Procedia Computer Science, vol. 167, pp. 254–262, 2020, doi: 10.1016/j.procs.2020.03.219.

[12] V. N. Dornadula and S. Geetha, "Credit Card Fraud Detection using Machine Learning Algorithms," Procedia Computer Science, vol. 165, pp. 631–641, 2019, doi: 10.1016/j.procs.2020.01.057.

[13] S. Carta, G. Fenu, D. Reforgiato Recupero, and R. Saia, "Fraud detection for E-commerce transactions by employing a prudential Multiple Consensus model," Journal of Information Security and Applications, vol. 46, pp. 13–22, Jun. 2019, doi: 10.1016/j.jisa.2019.02.007.

[14] S. Nami and M. Shajari, "Cost-sensitive payment card fraud detection based on dynamic random forest and k-nearest neighbors," Expert Systems with Applications, vol. 110, pp. 381–392, Nov. 2018, doi: 10.1016/j.eswa.2018.06.011.

[15] A. C. Bahnsen, A. Stojanovic, D. Aouada and B. Ottersten, "Cost Sensitive Credit Card Fraud Detection Using Bayes Minimum Risk," 2013 12th International Conference on Machine Learning and Applications, Miami, FL, 2013, pp. 333-338, doi: 10.1109/ICMLA.2013.68.

[16] U. Fiore, A. De Santis, F. Perla, P. Zanetti, and F. Palmieri, "Using generative adversarial networks for improving classification effectiveness in credit card fraud detection," Information Sciences, vol. 479, pp. 448–455, Apr. 2019, doi: 10.1016/j.ins.2017.12.030.

[17] N. F. Ryman-Tubb, P. Krause, and W. Garn, "How Artificial Intelligence and machine learning research impacts payment card fraud detection: A survey and industry benchmark," Engineering Applications of Artificial Intelligence, vol. 76, pp. 130–157, Nov. 2018.

[18] N. Japkowicz and S. Stephen, "The class imbalance problem: A systematic study," IDA, vol. 6, no. 5, pp. 429–449, Nov. 2002, doi: 10.3233/IDA-2002-6504.

[19] X. Zhang, Y. Han, W. Xu, and Q. Wang, "HOBA: A novel feature engineering methodology for credit card fraud detection with a deep learning architecture," Information Sciences, May 2019, doi: 10.1016/j.ins.2019.05.023.

[20] Roy, Abhimanyu & Sun, Jingyi & Mahoney, Robert & Alonzi, Loreto & Adams, Stephen & Beling, Peter. (2018). Deep learning detecting fraud in credit card transactions. 129-134. 10.1109/SIEDS.2018.8374722.

[21] Chouiekh, Alae & Haj, EL. (2018). ConvNets for Fraud Detection analysis. Procedia Computer Science. 127. 133-138. 10.1016/j.procs.2018.01.107.

[22] Wang, Deshen & Chen, Bintong & Chen, Jing. (2018). Credit Card Fraud Detection Strategies with Consumer Incentives. Omega. 88. 10.1016/j.omega.2018.07.001.

[23] F. Carcillo, A. Dal Pozzolo, Y.-A. Le Borgne, O. Caelen, Y. Mazzer, and G. Bontempi, "SCARFF: A scalable framework for streaming credit card fraud detection with spark," Information Fusion, vol. 41, pp. 182–194, May 2018, doi: 10.1016/j.inffus.2017.09.005.

[24] Rtayli, Naoufal & Enneya, Nourddine. (2020). Selection Features and Support Vector Machine for Credit Card Risk Identification. Procedia Manufacturing. 46. 941-948. 10.1016/j.promfg.2020.05.012.

[25] Y. Wu, Y. Xu, and J. Li, "Feature construction for fraudulent credit card cash-out detection," Decision Support Systems, vol. 127, p. 113155, Dec. 2019.

[26] M. F. Zeager, A. Sridhar, N. Fogal, S. Adams, D. E. Brown and P. A. Beling, "Adversarial learning in credit card fraud detection," 2017 Systems and Information Engineering Design Symposium (SIEDS), Charlottesville, VA, 2017, pp. 112-116, doi: 10.1109/SIEDS.2017.7937699.

[27] J. Jurgovsky et al., "Sequence classification for credit-card fraud detection," Expert Systems with Applications, vol. 100, pp. 234–245, Jun. 2018, doi: 10.1016/j.eswa.2018.01.037.

[28] H. Zhu, G. Liu, M. Zhou, Y. Xie, A. Abusorrah, and Q. Kang, "Optimizing Weighted Extreme Learning Machines for imbalanced classification and application to credit card fraud detection," Neurocomputing, vol. 407, pp. 50–62, Sep. 2020, doi: 10.1016/j.neucom.2020.04.078.

[29] S. Patil, V. Nemade, and P. K. Soni, "Predictive Modelling For Credit Card Fraud Detection Using Data Analytics," Procedia Computer Science, vol. 132, pp. 385–395, 2018.

244  Computer Science & Information Technology (CS & IT)

[30] A. Dal Pozzolo, G. Boracchi, O. Caelen, C. Alippi and G. Bontempi, "Credit Card Fraud Detection: A Realistic Modeling and a Novel Learning Strategy," in IEEE Transactions on Neural Networks and Learning Systems, vol. 29, no. 8, pp. 3784-3797, Aug. 2018, doi: 10.1109/TNNLS.2017.2736643.

## AUTHOR

**Yusuf Yazici** took his undergraduate degree from the Computer Engineering Department of Istanbul Technical University. He currently works as a Machine Learning and Artificial Intelligence Engineer at Turkcell, a leading digital telecommunication operator in Turkey. He also continues his master degree at the Computer Engineering Department of Istanbul Technical University.

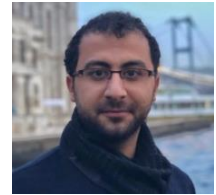

© 2020 By AIRCC Publishing Corporation. This article is published under the Creative Commons Attribution (CC BY) license.